\documentclass{article}
\usepackage{arxiv}

\usepackage{soul}
\usepackage[utf8]{inputenc} 
\usepackage[T1]{fontenc}    
\usepackage{hyperref}       
\usepackage{url}            
\usepackage{booktabs}       
\usepackage{amsfonts}       
\usepackage{nicefrac}       
\usepackage{microtype}      
\usepackage{lipsum}		
\usepackage{graphicx}
\usepackage{doi}
\usepackage{multirow}
\usepackage{multicol}
\usepackage{xcolor}
\usepackage{subcaption}
\usepackage[normalem]{ulem}
\usepackage{amsmath,amsfonts,bm}
\usepackage{amssymb,amsthm}
\usepackage{enumitem}
\usepackage{subcaption}
\usepackage{adjustbox}
\usepackage{afterpage}
\usepackage{xspace}
\usepackage[
    backend=biber,
    citestyle=authoryear,
    bibstyle=authortitle,
    mincitenames=1,
    maxcitenames=1,
    maxbibnames=3
]{biblatex}
\usepackage[group-separator={,}]{siunitx}
\usepackage{amsmath}
\usepackage{amsfonts}
\usepackage{makecell}
\usepackage{lipsum}
\usepackage{tcolorbox}
\usepackage{multirow}
\usepackage{marvosym}

\usepackage[capitalize]{cleveref}
\crefname{section}{Sec.}{Secs.}
\Crefname{section}{Section}{Sections}
\Crefname{table}{Table}{Tables}
\crefname{table}{Tab.}{Tabs.}
\Crefname{figure}{Figure}{Figures}
\crefname{figure}{Fig.}{Figs.}
\Crefname{equation}{Equation}{Equations}
\crefname{equation}{Eq.}{Eqs.}
\Crefname{algorithm}{Algorithm}{Algorithms}
\crefname{algorithm}{Alg.}{Algs.}

\usepackage[labelformat=simple]{subcaption}

\colorlet{lightpink}{pink!35}
\colorlet{lightcyan}{cyan!20}
\colorlet{lightgray}{gray!40}
\definecolor{darkgray}{rgb}{0.9, 0.9, 0.9}
\definecolor{lightgreen}{rgb}{0.886, 0.941, 0.851}
\colorlet{red}{red!80}
\colorlet{blue}{blue!80}
\colorlet{green}{green!60!black}
\colorlet{algemp}{cyan!10}
\colorlet{lightred}{red!50}

\usepackage{dirtree}
\usepackage{xcolor,colortbl}
\usepackage{pifont}
\usepackage{wrapfig}

\newcolumntype{a}{>{\columncolor{gray!20!white}}c}
\newcommand{\name}{{SimpleSeg}}

\newcommand{\citep}[1]{\parencite{#1}}
\renewcommand{\thefootnote}{\fnsymbol{footnote}}
\setlist[itemize,1]{leftmargin=\dimexpr 18pt}
\setlist[enumerate,1]{leftmargin=\dimexpr 18pt}

\title{
\raisebox{-0.1\height}{\includegraphics[width=0.032\textwidth]{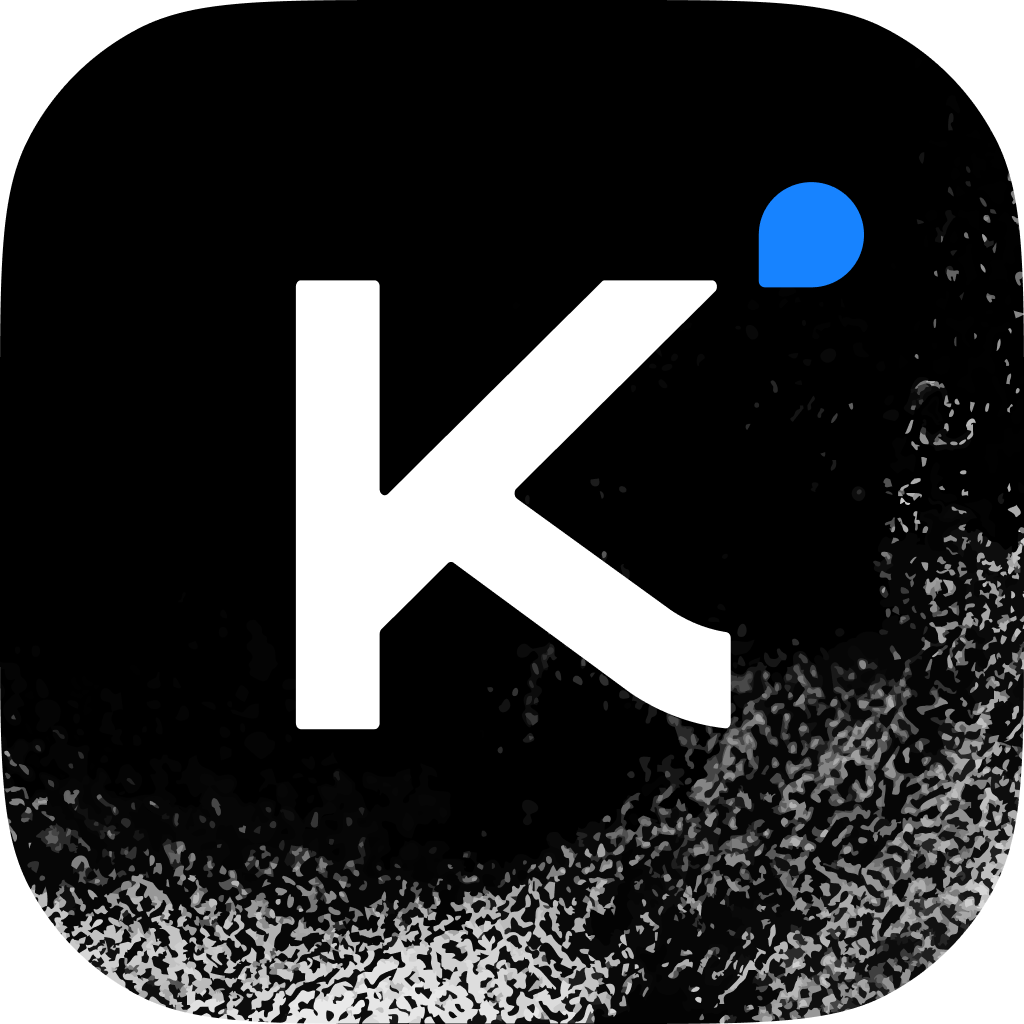}} %
Towards Pixel-Level VLM Perception \\ via Simple Points Prediction}

\author{
\\[-5mm]
\textbf{Tianhui Song}$^{1,2*\dag}$ \quad
\textbf{Haoyu Lu}$^{1*\ddag}$ \quad
\textbf{Hao Yang}$^{1}$ \quad
\textbf{Lin Sui}$^{1}$ \quad
\textbf{Haoning Wu}$^{1}$ \\
\textbf{Zaida Zhou}$^{1}$ \quad
\textbf{Zhiqi Huang}$^{1}$ \quad
\textbf{Yiping Bao}$^{1}$ \quad
\textbf{Y.Charles}$^{1}$ \quad
\textbf{Xinyu Zhou}$^{1}$ \quad
\textbf{Limin Wang}$^{2}$
\\[2ex]
$^1$ Moonshot AI \quad $^2$ Nanjing University
}

\date{}

\renewcommand{\headeright}{\name}
\renewcommand{\undertitle}{}
\renewcommand{\shorttitle}{\raisebox{-0.12\height}{\includegraphics[width=0.02\textwidth]{sections/figs/logo-new.png}}}

\renewcommand{\cite}{\citep}

\begin{document}
\maketitle

\let\thefootnote\relax\footnotetext{$^*$ Equal contribution.  $^\dag$ This work was done during interning at Moonshot AI.  $^\ddag$ Project lead.}

\vspace{-3mm}

\begin{abstract}
We present \textbf{\name}, a strikingly simple yet highly effective approach to endow Multimodal Large Language Models (MLLMs) with native pixel-level perception. 
Our method reframes segmentation as a simple sequence generation problem: the model directly predicts a \textbf{sequence of points} (textual coordinates) delineating object boundaries, entirely within its language space. 
To achieve high fidelity, we introduce a two-stage SFT→RL training pipeline, where Reinforcement Learning with an IoU-based reward refines the point sequences to accurately match ground-truth contours. 
We find that \emph{\textbf{the standard MLLM architecture possesses a strong, inherent capacity for low-level perception}} that can be unlocked without any specialized architecture. 
On segmentation benchmarks, \name~achieves performance that is comparable to, and often surpasses, methods relying on complex, task-specific designs. 
This work lays out that precise spatial understanding can emerge from simple point prediction, challenging the prevailing need for auxiliary components and paving the way for more unified and capable MLLMs. Code, data, and model are publicly accessible at \url{https://simpleseg.github.io/}. 

\end{abstract}

\begin{figure}[htb]
    \centering
    \includegraphics[width=0.82\linewidth]{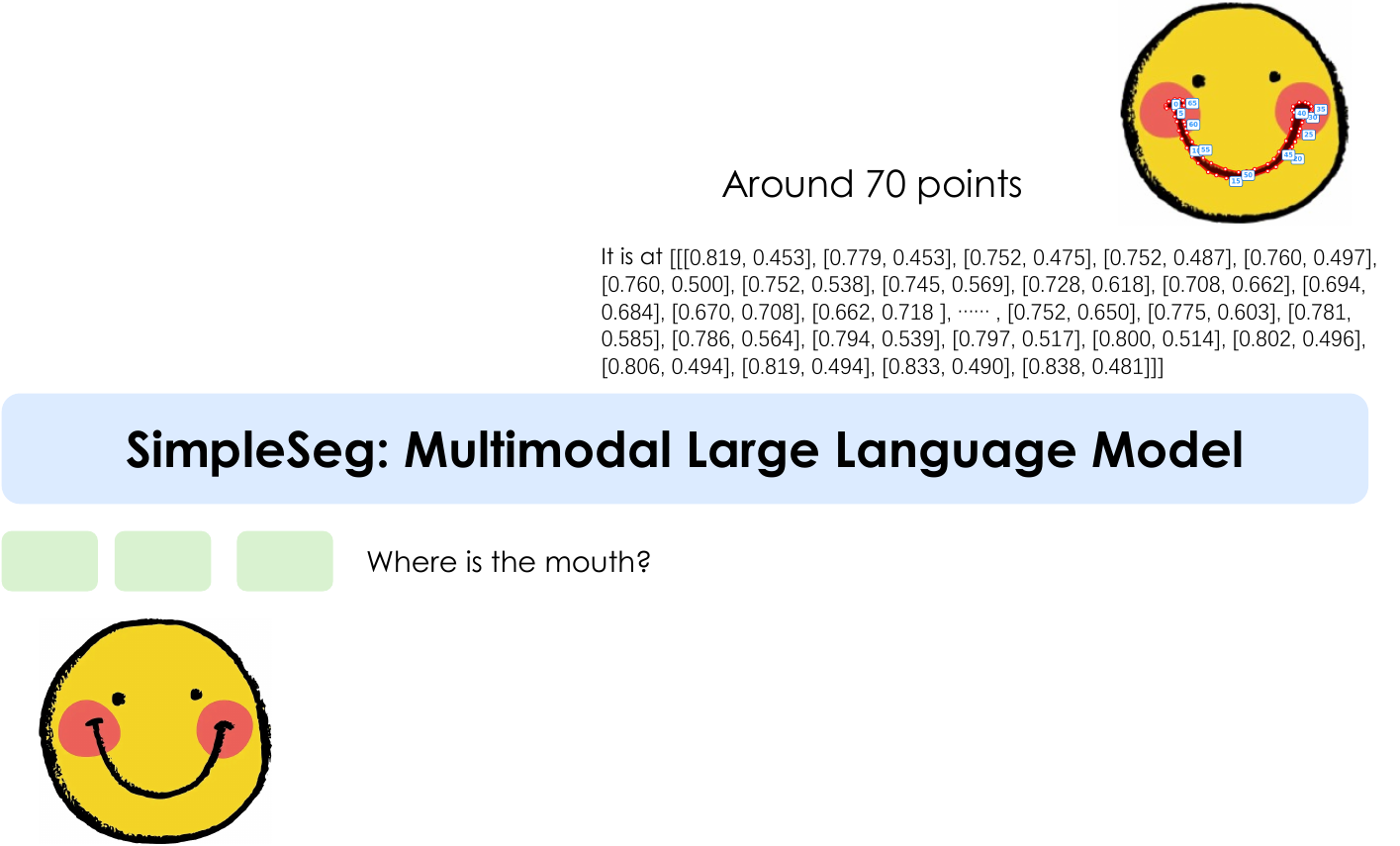}
    \caption{In this work, we explore the limits of MLLM pixel-level perception by predicting the next point in a contour with the simplest approach possible. Without introducing any complex architectures or special patterns, we show how even minimalistic point prediction can achieve effective segmentation at the pixel level.}
    \label{fig:samples}
\end{figure}

\section{Introduction}

Multimodal Large Language Models (MLLMs) have rapidly advanced open-ended vision--language understanding, delivering strong performance across captioning, VQA, and interactive grounding~\cite{liu2024llava,openai2024gpt4ocard,comanici2025gemini}.

Yet, despite impressive semantic competence, today’s Multimodal Large Language Models (MLLMs) remain largely \emph{image-level} in their perception, struggling to precisely localize and delineate fine structures—from object boundaries to thin parts—that are essential for genuine spatial understanding.
This limitation is partly rooted in the evolution of multimodal foundational models.
While perception is a cornerstone of multimodal tasks, dense prediction tasks like segmentation have historically been overlooked as a foundational capability, as they often rely on specialized decoders or complex architectural designs not native to language-centric models.
In contrast, object grounding and detection have been widely adopted, largely because bounding boxes can be conveniently represented as plain text coordinates (e.g., $x_1, y_1, x_2, y_2$) and easily integrated into the pre-training pipeline.
However, we argue that the coarse localization offered by bounding boxes is insufficient for the next generation of applications.
Such pixel-level grounding is not merely cosmetic: it is foundational for controllable image editing~\cite{shi2024seededit}, vision-based tool use~\cite{wang2025mllm}, and GUI-grounded agents~\cite{liu2025infiguiagent,qin2025ui} that must reason, act, and communicate about precisely \emph{where} things are.
Therefore, to bridge this gap, we move beyond coarse bounding boxes and consider a more precise and granular approach: point prediction.

\begin{figure}[h]
    \centering
    \vspace{-2mm}
    \includegraphics[width=0.999\linewidth]{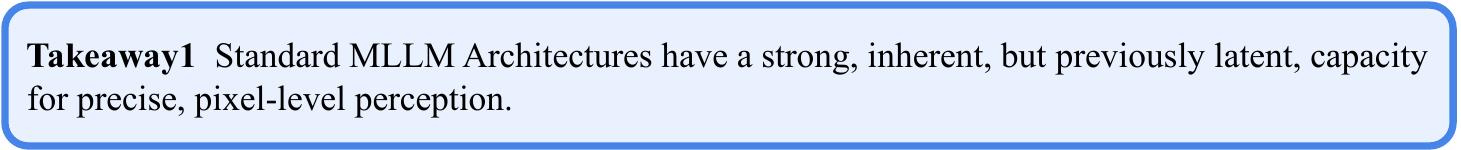}
    \label{fig:tw1}
    \vspace{-5mm}
\end{figure}

A prevalent approach augments MLLMs with task-specific decoders (e.g., SAM- or RPN-style heads) on top of the multimodal backbone~\cite{lai2024lisa,zhang2024groundhog,he2024multi,ren2024pixellm,rasheed2024glamm,zhang2023next,wu2024towards,wu2024visionllmv2}. While effective, this design couples architecture to specific tasks, complicates end-to-end training with extra parameters, and pushes outputs out of the language space, weakening interpretability and compositional reasoning. As a result, fine-grained perception remains underexplored as a core capability of native MLLMs. Decoder-free methods, such as Text4Seg~\cite{lan2024text4seg}, serialize masks as text, but suffer from dense token budgets and compromised interpretability. VisionLLM~\cite{wang2024visionllm} emits polygons but restricts them to a small number of vertices, limiting its performance. Both methods fail to deliver pixel-level segmentation with the generality and reasoning fluency of modern MLLMs.

\begin{figure}[t!]
    \centering
    \includegraphics[width=0.999\linewidth]{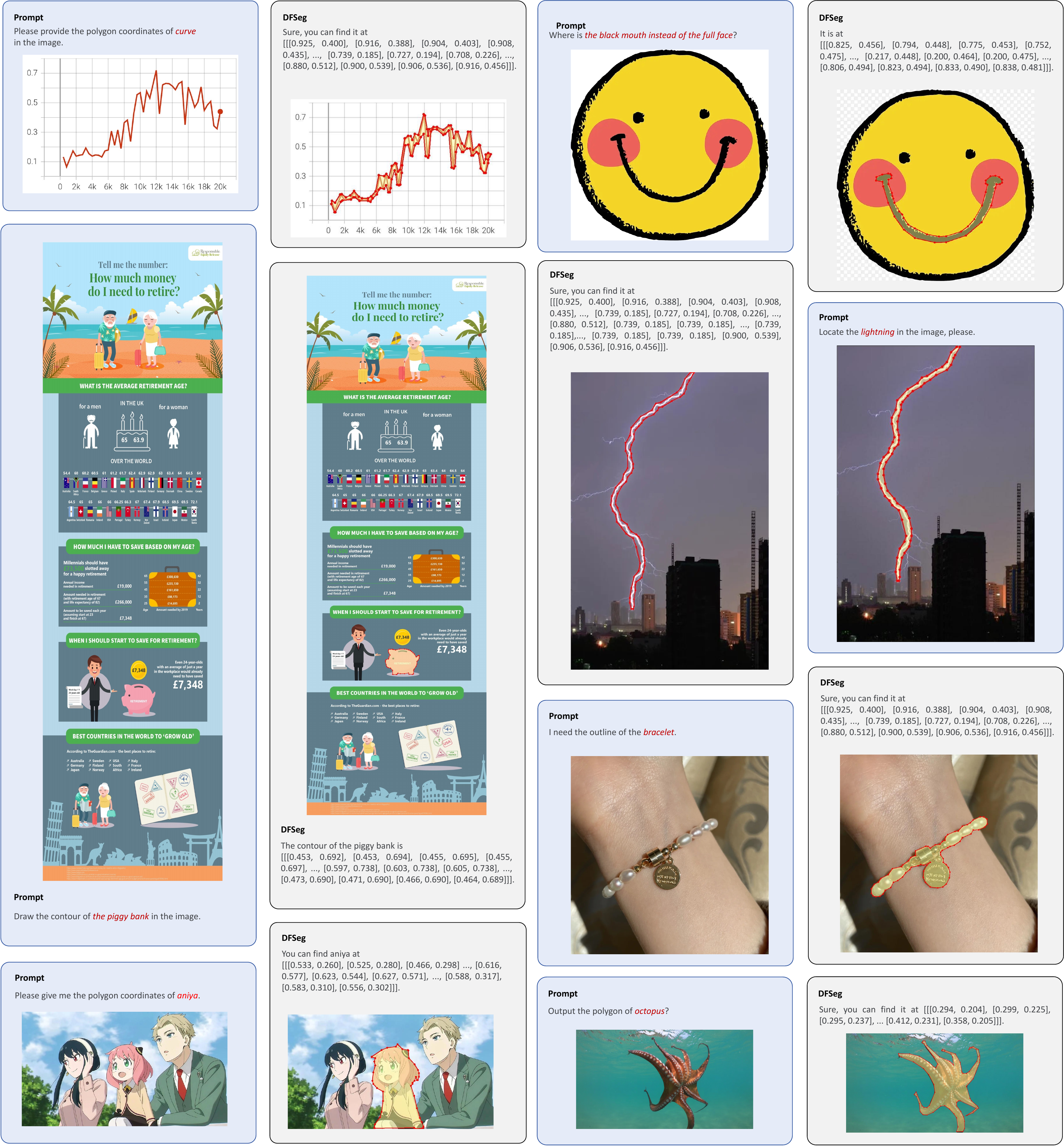}
    \caption{Segmentation results of SimpleSeg on natural and non-natural images. These examples highlight the model's excellent generalization, showing its precise pixel-level perception is not confined to real-world objects. The model successfully segments targets from natural photographs (the lightning) and performs with equal precision on various forms of "in-screen" or digitally generated content, including anime, data charts, and infographics.}
    \label{fig:samples}
\end{figure}

\begin{figure}[t!]
    \centering
    \includegraphics[width=0.999\linewidth]{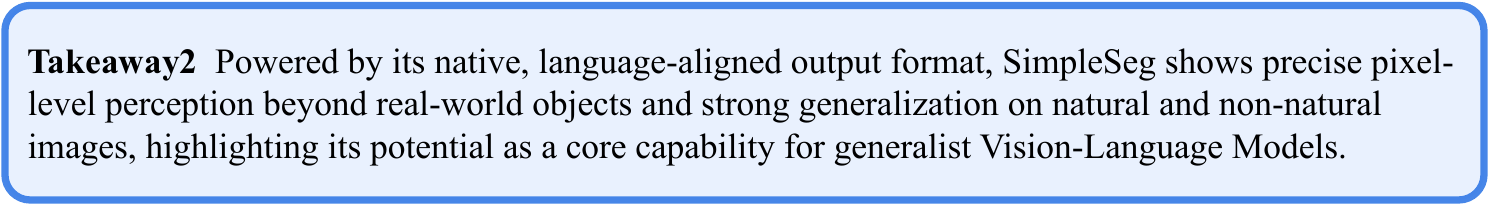}
    \label{fig:tw2}
    \vspace{-1cm}
\end{figure}

In this work, we investigate a strikingly simple question: can an MLLM achieve high-fidelity segmentation by merely predicting a sequence of points? We present \textbf{\name}, a minimalist decoder-free approach that reframes segmentation as simple, sequential point prediction entirely within the language space. More than just a method, our work serves as a crucial finding: \emph{\textbf{we demonstrate that standard MLLM architectures possess a strong, inherent capacity for fine-grained perception}}, a potential that can be unlocked without any specialized decoders or complex output formats. This approach preserves the model's generalist architecture, dramatically simplifies the training pipeline, and naturally unifies object localization tasks (points, boxes, and masks) under a single, human-readable textual interface.

Specifically, we first introduce a systematic point-sequence-based representation for segmentation masks that efficiently scales data preparation. Based on this, we generalize the perceptual localization task beyond text queries: any target can be an input or output in a 4-tuple, \([text, point, box, mask]\), allowing for a rich combination of task formats that boosts data efficiency and robustness.

To make this simple point prediction effective, we design a two-stage SFT→RL training pipeline. After a standard supervised fine-tuning (SFT) stage to learn the basic task format, we pioneer the use of Reinforcement Learning (RL) to optimize the entire generated sequence of points. By using an IoU-based reward, RL directly refines the fidelity and closure of the resulting shape without altering the MLLM's architecture. To our knowledge, this is the first work to successfully apply reinforcement learning to a decoder-free MLLM for segmentation.

Empirically, our model attains high-quality, native fine-grained perception and generalizes robustly across diverse domains and resolutions, as illustrated in \cref{fig:samples}. On challenging referring benchmarks such as the refCOCO series, \name~achieves performance that is comparable to, and often surpasses, prominent methods that rely on complex, task-specific decoders.
The main contributions can be summarized as follows:
\begin{itemize}
\item We present a minimalist, decoder-free approach for MLLM segmentation based on simple point sequence prediction, challenging the necessity of complex architectural additions.
\item We provide a key finding that standard MLLM architectures possess a strong, inherent potential for pixel-level perception, which can be unlocked with the right training methodology.
\item We are the first to propose and validate an SFT→RL pipeline for this task, using sequence-level IoU rewards to directly optimize the quality of the generated geometry.
\item We demonstrate that our simple approach achieves performance comparable to or exceeding that of more complex decoder-based systems on standard referring segmentation benchmarks.
\end{itemize}

In addition to the above, \name~offers several key benefits:
\begin{itemize}
    \item \textbf{Simplicity}: \name~requires no specialized modules and adheres to the standard MLLM architecture, it can be seamlessly and efficiently integrated as a new, core pre-training task for foundation models, similar to visual grounding.
    \item \textbf{Task Generality}: By framing segmentation as a text-generation problem, our approach is inherently flexible. The model can be easily adapted to a wide range of vision-language tasks that require precise spatial localization.
    \item \textbf{Interpretable Output}: The model generates explicit, human-readable coordinate sequences instead of dense pixel masks. This transparency simplifies debugging and makes the output directly usable for downstream applications like interactive editing or tool use.
\end{itemize}

This makes \name~not just an efficient solution, but a versatile framework for deploying pixel-level perception in multimodal models with a broad range of applications.

\section{Related Work}

\paragraph{Multimodal Large Language Models.}
Multimodal Large Language Models (MLLMs) have significantly advanced vision-language tasks by extending the reasoning capabilities of LLMs to the visual domain~\cite{yin2024survey}. Early pioneering models like LLaVA~\cite{liu2024llava} established a strong foundation for multimodal instruction following. Subsequent works~\cite{lu2024deepseek, openai2024gpt4ocard, comanici2025gemini} have further pushed the boundaries of performance by scaling up model and data size. However, a common limitation persists: the perception of these models is typically coarse and image-level. They excel at high-level description and reasoning but lack the native ability for precise, pixel-level localization, which remains a largely underexplored frontier.

\paragraph{Approaches to Pixel-Level MLLM Perception.}
Efforts to equip MLLMs with dense, pixel-level perception have primarily followed two distinct paths, creating a central dilemma between performance and architectural integrity. The first, a \textbf{hybrid approach}, augments a general MLLM backbone with specialized, task-specific decoders~\cite{lai2024lisa,zhang2024groundhog,ren2024pixellm,wu2024visionllmv2,rasheed2024glamm,xia2024gsva,zhang2024omg}. These external modules, often inspired by SAM or other segmentation architectures, achieve strong performance on specific tasks. However, this modular design compromises architectural purity; it introduces extra parameters, complicates training, and moves the final output outside the native language space.
GiT~\cite{wang2024git} uses a Vision Transformer (ViT) backbone and tokenizes text input via simple word embeddings to achieve open-vocabulary segmentation. It relies on an architecture customized for visual tasks, limiting the generalization power compared to a true vision-language model.

The second are \textbf{unified approaches}. Some works~\cite{lan2024text4seg,wang2024visionllm,chen2022unified,chen2021pix2seq} attempt to keep all outputs strictly within the language space by representing masks as text sequences, e.g., RLE~\cite{lan2024text4seg} or polygons~\cite{wang2024visionllm,chen2022unified}. While conceptually aligned with the end-to-end philosophy of LLMs, these methods have struggled with fidelity, suffering from excessive token consumption and an inability to capture fine-grained details. UFO~\cite{tang2025ufo} proposes an extra, special mask token and relies on hacking the intermediate feature for retrieval mechanisms. Consequently, achieving high-performance, pixel-level perception \emph{natively} within a standard MLLM architecture—without sacrificing simplicity, resolution, or generalization—remains a fundamental open challenge.  
\section{Methodology}
We present \textbf{\name}, a simple yet effective framework that equips a vanilla MLLM with native pixel-level perception \emph{via simple points prediction}.
The key idea is to keep segmentation entirely inside the language space by predicting a \emph{point trajectory} (i.e., an explicit sequence of 2D coordinates) that traces the target contour.
This design is decoder-free, architecture-agnostic, and naturally unifies points, boxes, and masks under one textual interface.

\subsection{Task Formulation and Data Construction}
\label{subsec:data}

\begin{figure}[t!]
    \centering
    \includegraphics[width=0.999\linewidth]{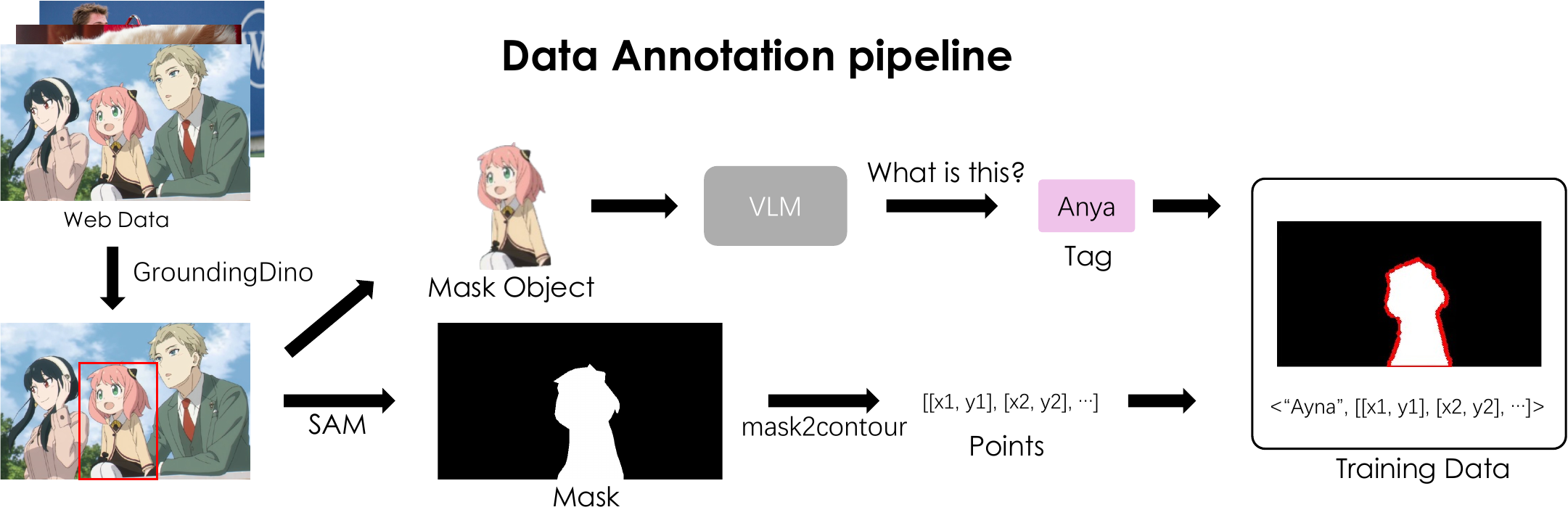}
    \caption{Overview of our data annotation pipeline, which incorporates modules for object detection, mask segmentation, points conversion, and instance caption.}
    \label{fig:datapp}
\end{figure}

\paragraph{Outputs as Text in One Space.}
Perceptual location information commonly appears as (i) a center \texttt{point}, (ii) a \texttt{bbox}, or (iii) a \texttt{mask}.
Keeping all outputs as \emph{text tokens} preserves the MLLM's generalist interface and enables direct composition with language prompts and tools.
We therefore adopt the following normalized, human-aligned formats:
\begin{align*}
\texttt{$<$point$>$} &:~ [x, y] \\
\texttt{$<$bbox$>$}  &:~ [x_1, y_1, x_2, y_2] \\
\texttt{$<$mask$>$}  &:~ [[x_1, y_1], [x_2, y_2], \dots, [x_{V}, y_{V}]]
\end{align*}
where coordinates are normalized to $[0,1]$ and $V$ is variable.

\paragraph{Masks as Point Trajectories (Contours).}
Instead of dense per-pixel encodings (e.g., R-RLE~\cite{lan2024text4seg}), we represent a mask by an \emph{explicit point trajectory} that sparsely samples its boundary.
This brings three benefits: (1) \textbf{interpretability} (human-readable coordinates), (2) \textbf{compositionality} (same token space as text/points/boxes), and (3) \textbf{controllable token budget} (linear in the number of vertices rather than image resolution).
For training data derived from binary masks, we extract polygonal contours using the Suzuki–Abe algorithm~\cite{suzuki1985topological} (OpenCV~\cite{itseez2015opencv}), enforce a consistent traversal order (clockwise), and optionally apply a tolerance-based sparsification to obtain a compact simple points sequence.

\paragraph{A Unified Query Interface.}
Following the spirit of promptable segmentation, we model grounding over the tuple
\[
\texttt{target}=[\texttt{text}, \texttt{point}, \texttt{bbox}, \texttt{mask}],
\]
and instantiate \emph{queries} as Cartesian products of the available elements (e.g., \texttt{(text$\!\to$bbox)}, \texttt{(point$\!\to$mask)}).
Two examples are:

\texttt{\textbf{Q:} What is the bounding box of $<$text$>$?~~ \textbf{A:} $<$bbox$>$.}

\texttt{\textbf{Q:} Give the polygon of the object at $<$point$>$?~~ \textbf{A:} $<$mask$>$.}

This interface (i) multiplies supervision sources by recombining weak labels (e.g., points/boxes from masks via min/max or centroid), and (ii) standardizes outputs for instruction tuning and RL.

\paragraph{Text Grammar.}
We constrain outputs with a minimal JSON-like grammar to reduce decoding entropy, and they can be parsed automatically at inference:
\[
\underbrace{\texttt{[[x, y], [x, y], \dots]}}_{\text{polygon}},~
\underbrace{\texttt{[x, y]}}_{\text{point}},~
\underbrace{\texttt{[x1, y1, x2, y2]}}_{\text{bbox}}.
\]

\paragraph{Data Annotation Pipeline.}
To scale our framework with large-scale web data, we construct an automatic data annotation pipeline, as shown in \cref{fig:datapp}, to generate instance-level segmentation labels.
Specifically, the pipeline employs: Grounding-DINO~\cite{liu2023grounding} for phrase grounding and object detection, SAM to extract segmentation masks, the algorithm for converting mask to contour coordinates, and an off-the-shelf VLM for optional, refined object description tagging.

\subsection{Training Pipeline of \name}
\label{subsec:train}
Our training including: (i) instruction tuning (SFT) to cold-start structured generation, and (ii) reinforcement learning (RL) to optimize sequence-level, location-aware objectives.

\paragraph{Stage I: Instruction Tuning.}
According to the aforementioned polygon-based representation of masks, we curate instruction–response pairs spanning
\texttt{(text$\!\leftrightarrow$point)}, \texttt{(text$\!\leftrightarrow$bbox)}, and \texttt{(text/point$\!\to$mask)}.
The supervised finetuning stage aims to teach the MLLM to emit correct output formats, including well-formed coordinates, closing brackets, and consistent ordering, while learning basic grounding priors.
Empirically, this already yields competitive performance and provides a stable initialization for RL.

\paragraph{Stage II: Reinforcement Learning with GSPO.}
Reinforcement learning (RL) has demonstrated significant effectiveness in reasoning tasks for MLLM~\cite{shao2024deepseekmath,guo2025deepseek,team2025kimik15}.
However, the potential of RL for sharpening fine-grained perception remains largely untapped. 
While SFT aligns tokens to local supervision, pixel-level segmentation quality depends on global properties of the entire sequence (closure, boundary fidelity, and verbosity).
We focus on leveraging RL to boost the perception accuracy of MLLM, since, in essence, reinforcement learning is a more reasonable and efficient optimization method for perception tasks.
Especially under our data and task formulation, we are not aiming to force the model to rigidly regress fixed ground-truth coordinates in the training data, as contour sequences are inherently flexible.
The optimization process relies more on location-aware rewards to explore and refine predictions at the sequence level, while format-rule-based judges can enforce valid, parseable output structures.
We therefore adopt GSPO~\cite{zheng2025group} as our RL algorithm, and adopt a rule-based reward system that mainly consists of three types of rewards:
\begin{itemize}
   \item \textbf{Mask IoU} reward: The direct IoU between the predicted and ground-truth mask. The range of reward values is $[0.0, 0.1]$. We set a threshold $\tau$ that the reward is $0$ if IoU is less than $\tau$.
   \item \textbf{MSE Distance IoU} reward: The negative mean square distance between the centroids of predicted and ground-truth mask. It is normalized with the image size.
    \item \textbf{Format} reward: In addition to the accuracy reward model, we employ a format reward that enforces the model to output correct polygon coordinates formats. If the format is wrong, the reward returns zero.
\end{itemize}
Crucially, RL lets the model discover alternative yet valid trajectories (e.g., different starting points, equivalent vertex sets) instead of overfitting to a single annotation.

\paragraph{Why RL for Point Trajectories?}
As far as we know, we are the first to leverage RL in the realm of decoder-free MLLM segmentation.
Contours are inherently many-to-one w.r.t.\ masks; enforcing exact token matching is suboptimal.
Reinforcement learning well bridges the gap and evaluates the \emph{rendered mask}, directly aligning optimization with the end metric, and improves closure and thin-structure adherence that are difficult to teach via token-level losses alone.






\section{Experiment}

\subsection{Implementation Details}

We validate our method on two open-source MLLM architectures, \emph{Qwen2.5-VL-7B}~\cite{Qwen-VL} and \emph{Kimi-VL}~\cite{team2025kimivl}, an efficient MoE model with 2.8B activated parameters.
Training uses 32 NVIDIA GPUs with a global batch size of 256 and the enhanced Muon optimizer~\cite{liu2025muon}.
For supervised fine-tuning (SFT), we use an initial learning rate of \(5{\times}10^{-5}\) with cosine decay to \(2{\times}10^{-6}\), and a warm-up ratio of 0.03.
For reinforcement learning (RL), we adopt GSPO with clip ratio in \([3{\times}10^{-4},4{\times}10^{-4}]\) and a KL coefficient of 0.01.
Unless otherwise specified, coordinates are normalized and serialized in the text space using our polygon format, and they are sparsified by a tolerance parameter \(\epsilon\) (cf. \cref{sec:explor}). 
More implementation details are provided in the Appendices.

\vspace{-0.1cm}
\subsection{Main Results}

\paragraph{Referring Expression Segmentation}
The referring expression segmentation (RES) task aims to segment the object in an image that is described by a given natural-language expression.
We follow the training recipe of \cite{lan2024text4seg}, which constructs the training dataset with the \texttt{train} split of refCOCO, refCOCO+~\cite{kazemzadeh2014referitgame}, refCOCOg~\cite{mao2016generation}, and refCLEF.
As shown in \cref{tab:results_res}, our \name~achieves superior performance in decoder-free models, and comparable to decoder-based methods.
This demonstrates our method’s strong fine-grained perception capacity as a generalist vision-language model, without any modification of model architecture.

\paragraph{Referring Expression Comprehension}
Our \name~is also directly usable for object detection by converting the predicted mask to a bounding box through simple min-max operations.
Accordingly, we evaluate our approach on the Referring Expression Comprehension (REC) task, using the same model trained as in RES.
For the evaluation specification, we calculate the average accuracy with a threshold IoU of 0.5 between the predicted and ground truth bounding boxes.
As shown in \cref{tab:results_rec}, our \name~obtains state-of-the-art performance on the benchmarks.
Specifically, our method achieves an average score of 87.2, exceeding the closest competitor, Text4Seg, even though it is equipped with a mask refiner.

\begin{figure}[h]
    \centering
    \includegraphics[width=0.999\linewidth]{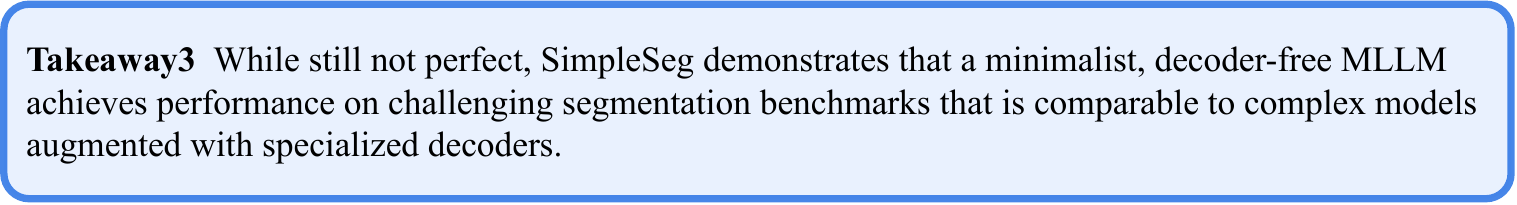}
    \label{fig:tw3}
    \vspace{-1cm}
\end{figure}
\vspace{2mm}

\begin{table}[t]
\caption{\textbf{Referring Expression Segmentation} results (cIoU) on refCOCO (+/g) datasets \cite{kazemzadeh2014referitgame,mao2016generation}, compared to approaches that adopt MLLMs for segmentation. $^*$ denotes the model underwent pre-training.} 
\label{tab:results_res}
\centering
\resizebox{\linewidth}{!}{
\begin{tabular}{l|ccc|ccc|cc|c}
    \toprule
    \multirow{2}*{Methods} & \multicolumn{3}{c|}{refCOCO} & \multicolumn{3}{c|}{refCOCO+} & \multicolumn{2}{c|}{refCOCOg} & \multirow{2}*{Avg.}\\
    \cline{2-9}
     & val & testA & testB & val & testA & testB & val & test & \\
    \bottomrule
    \multicolumn{10}{c}{\textit{Decoder-based Models}} \\
    NEXT-Chat \cite{zhang2023next} & 74.7 & 78.9 & 69.5 & 65.1 & 71.9 & 56.7 & 67.0 & 67.0 & 68.9 \\
    LISA \cite{lai2024lisa} & 74.9 & 79.1 & 72.3 & 65.1 & 70.8 & 58.1 & 67.9 & 70.6 & 69.9 \\
    PixelLM \cite{ren2024pixellm} & 73.0 & 76.5 & 68.2 & 66.3 & 71.7 & 58.3 & 69.3 & 70.5 & 69.2 \\
    AnyRef \cite{he2024multi} & 76.9 & 79.9 & 74.2 & 70.3 & 73.5 & 61.8 & 70.0 & 70.7 & 72.2 \\
    GSVA \cite{xia2024gsva} & 77.2 & 78.9 & 73.5 & 65.9 & 69.6 & 59.8 & 72.7 & 73.3 & 71.4 \\
    LaSagnA \cite{wei2024lasagna} & 76.8 & 78.7 & 73.8 & 66.4 & 70.6 & 60.1 & 70.6 & 71.9 & 71.1 \\ 
    Groundhog \cite{zhang2024groundhog} & 78.5 & 79.9 & 75.7 & 70.5 & 75.0 & 64.9 & 74.1 & 74.6 & 74.2 \\ 
    Text4Seg (w/ SAM) & 79.2 & 81.7 & 75.6 & 72.8 & 77.9 & 66.5 & 74.0 & 75.3 & 75.4 \\
    \bottomrule
    \multicolumn{10}{c}{\textit{Decoder-free  Models}} \\
    UFO$_\text{LLaVA-1.5-7B}$~\cite{tang2025ufo} & 77.2 & 79.4 & 73.8 & 70.8 & 75.5 & 64.1 & 72.1 & 73.2 & 73.3 \\
    Text4Seg$_{\text{InternVL2-8B}}$~\cite{lan2024text4seg} & 74.7 & 77.4 & 71.6 & 68.5 & 73.6 & 62.9 & 70.7 & 71.6 & 71.4 \\
    \textbf{\name}$_\text{Qwen2.5-VL-7B}$ & 75.6 & 78.7 & 72.0 & 68.9 & 74.4 & 62.3 & 70.8 & 72.5 & 71.9 \\
    \textbf{\name}$_\text{Kimi-VL}$ & 76.9 & 78.9 & 73.6 & 71.1 & 75.2 & 66.1 & 72.8 & 74.3 & 73.6 \\
    \textbf{\name}$_\text{Qwen2.5-VL-7B}^*$ & 80.9 & 77.8 & 75.2 & 72.4 & 77.3 & 66.1 & 73.3 & 74.1 & 74.6 \\
    \textbf{\name}$_\text{Kimi-VL}^*$ & 80.0 & 80.6 & 76.2 & 70.4 & 76.2 & 67.1 & 72.8 & 74.7 & 74.8 \\

    \bottomrule
\end{tabular}
\textbf{}}
\end{table}

\begin{table}[t]
\caption{\textbf{Referring Expression Comprehension} results (Acc@0.5) on RefCOCO (+/g) datasets, compared to approaches that adopt MLLMs for segmentation. $^*$ denotes the model underwent pre-training.}
\label{tab:results_rec}
\centering
\resizebox{\linewidth}{!}{
\begin{tabular}{l|ccc|ccc|cc|c}
\toprule
\multirow{2}*{Methods} & \multicolumn{3}{c|}{refCOCO} & \multicolumn{3}{c|}{refCOCO+} & \multicolumn{2}{c|}{refCOCOg} & \multirow{2}*{Avg.}\\
\cline{2-9}
 & val & testA & testB & val & testA & testB & val & test & \\
\bottomrule
\multicolumn{10}{c}{\textit{Decoder-based Models}} \\

LISA \cite{lai2024lisa} & 85.4 & 88.8 & 82.6 & 74.2 & 79.5 & 68.4 & 79.3 & 80.4 & 79.8 \\
GSVA \cite{xia2024gsva} & 86.3 & 89.2 & 83.8 & 72.8 & 78.8 & 68.0 & 81.6 & 81.8 & 80.3 \\
NEXT-Chat \cite{zhang2023next} & 85.5 & 90.0 & 77.9 & 77.2 & 84.5 & 68.0 & 80.1 & 79.8 & 80.4\\
PixelLM \cite{ren2024pixellm} & 89.8 & 92.2 & 86.4 & 83.2 & 87.0 & 78.9 & 84.6 & 86.0 & 86.0\\
Text4Seg (w/ SAM) & 90.3 & 93.4 & 87.5 & 85.2 & 89.9 & 79.5 & 85.4 & 85.4 & 87.1\\
\bottomrule
\multicolumn{10}{c}{\textit{Decoder-free Models}} \\
UFO$_\text{LLaVA-1.5-7B}$~\cite{tang2025ufo} & 90.8 & 93.0 & 87.3 & 85.5 & 90.5 & 78.6 & 86.9 & 87.2 & 87.5 \\
Text4Seg$_{\text{InternVL2-8B}}$~\cite{lan2024text4seg} & 88.3 & 91.4 & 85.8 & 83.5 & 88.2 & 77.9 & 82.4 & 82.5 & 85.0 \\
\textbf{\name}$_\text{Qwen2.5-VL-7B}$ & 89.4 & 92.7 & 84.8 & 82.9 & 87.9 & 76.6 & 82.5 & 84.7 & 85.2 \\
\textbf{\name}$_\text{Kimi-VL}$ & 90.5 & 92.9 & 86.8 & 85.3 & 89.5 & 80.2 & 86.1 & 86.5 & 87.2 \\
\textbf{\name}$_\text{Qwen2.5-VL-7B}^*$ & 90.2 & 92.9 & 86.1 & 84.6 & 90.5 & 79.0 & 84.9 & 85.6 & 86.7 \\
\textbf{\name}$_\text{Kimi-VL}^{*}$ & 91.3 & 92.1 & 87.1 & 82.6 & 88.3 & 79.3 & 84.6 & 86.3 & 86.5 \\
\bottomrule
\end{tabular}
}
\end{table}

\begin{table}[t]
    \centering
    \caption{The gIoU score with different training stages in validation sets.}
    \begin{tabular}{ccc|ccc}
        \toprule
        Pre-train & SFT & RL & refCOCO & refCOCO+ & refCOCOg \\
        \midrule
         & $\checkmark$ & & 65.5 & 60.8 & 60.4 \\
         & $\checkmark$ & $\checkmark$ & 75.2 \color{green}{($\uparrow$ 9.7)} & \textbf{70.6} \color{green}{($\uparrow$ 9.8)} & 70.9 \color{green}{($\uparrow$ 10.5)} \\
        \midrule
        $\checkmark$ & & & 25.3 \color{red}{($\downarrow$ -45.7)}& 18.7 \color{red}{($\downarrow$ -46.4)} & 25.7 \color{red}{($\downarrow$ -43.0)}\\
        $\checkmark$ & $\checkmark$ & & 70.1 \color{green}{($\uparrow$ 4.6)} & 65.0 \color{green}{($\uparrow$ 4.2)} & 65.7  \color{green}{($\uparrow$ 5.3)}\\
        $\checkmark$ & $\checkmark$ & $\checkmark$ & \textbf{78.5} \color{green}{($\uparrow$ 13.0)} & 69.8 \color{green}{($\uparrow$ 9.0)} & \textbf{71.7} \color{green}{($\uparrow$ 11.3)} \\
        \bottomrule
    \end{tabular}
    \label{tab:abl-stage}
    \vspace{-2mm}
\end{table}

\begin{wrapfigure}{r}{0.48\textwidth}
\label{fig:abl-epsilon}
\centering
\includegraphics[width=\linewidth]{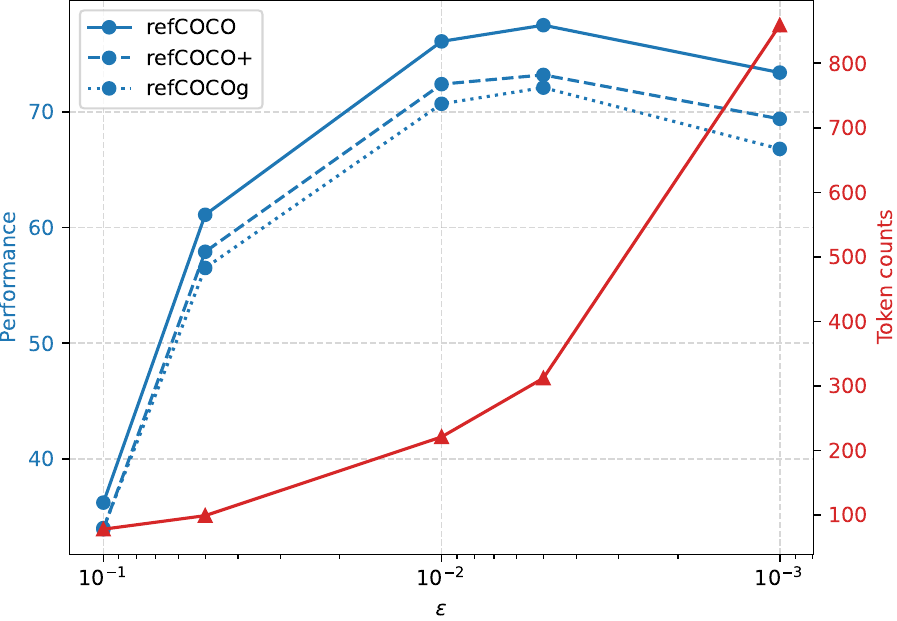}
\caption{The relationship between the sequence length and performance under the control of the point density parameter $\epsilon$.}
\label{fig:abl-epsilon}
\vspace{-2mm}
\end{wrapfigure}

\newpage
\subsection{Exploration Studies}
\label{sec:explor}

\paragraph{Effect of Training Stages}
Table~\ref{tab:abl-stage} ablates on the effect of different training stages, including pre-training, SFT, and RL.
We evaluate the gIoU score on the validation set of different datasets.
SFT alone reaches \textbf{65.5}, \textbf{60.8}, and \textbf{60.4} gIoU on three datasets, establishing a strong baseline from purely supervised polygon learning.
Adding RL lifts performance to \textbf{75.2} ($+9.7$), \textbf{70.6} ($+9.8$), and \textbf{70.9} ($+10.5$) respectively by a large margin, indicating that sequence-level credit assignment with IoU-based rewards is important for accurate \emph{closed} polygon generation and token-economical outputs.
Pre-training without SFT performs poorly (25.3 gIoU), and this stems from a distribution shift between pre-training and SFT prompts.
Pre-training lacks RefCOCO-style questions, and this phase primarily focuses on utilizing weakly labeled data to establish the model's basic segmentation ability.
It can be seen that both SFT and SFT+RL benefit significantly from pre-training, respectively, raising the gIoU by \textbf{4.6} and \textbf{13.0} on refCOCO, confirming that scaling the training data strengthens perceptual priors and benefits downstream tasks.

\begin{figure}[h]
    \centering
    \includegraphics[width=0.999\linewidth]{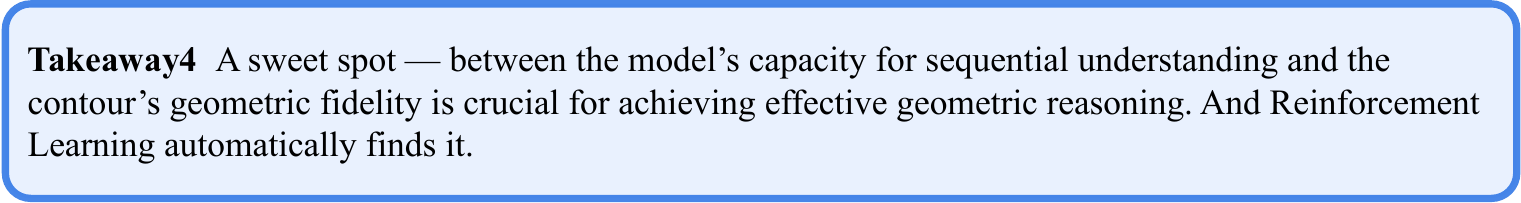}
    \label{fig:tw4}
    \vspace{-0.6cm}
\end{figure}

\paragraph{Point/Polygon Density (\(\epsilon\))}
$\epsilon$ is a hyperparameter that controls the polygon approximation accuracy for mask-to-contour conversion.
A smaller $\epsilon$ yields higher polygon precision and thus more sampled points.
Fig~\ref{fig:abl-epsilon} varies the sparsification tolerance.
We conduct SFT experiments with different $\epsilon$ and the results are demonstrated in \cref{fig:abl-epsilon}.
Performance is \emph{unimodal} w.r.t. token length: too few points underfit shapes (35.6 cIoU at 78 tokens), too many induce long-horizon decoding errors and length exposure (72.5 cIoU at 859 tokens), while a moderate density (221 tokens) yields the best score.

\begin{wrapfigure}{l}{0.48\textwidth}
\centering

\caption{gIoU score with different rewards.}
\begin{tabular}{c|ccc}
\toprule
Reward & RefC & RefC+ & RefCg\\
\midrule
 IoU & 76.9 & 71.9 & 72.9 \\
 + Distance & 77.1 & 72.2 & 73.1 \\
 + Length Penalty & 66.7 & 62.4 & 62.0 \\
\bottomrule
\end{tabular}
\label{tab:abl-reward}
\end{wrapfigure}

\paragraph{Reward Design for RL}
Table~\ref{tab:abl-reward} studies the effectiveness of different reward components, including distance and length penalty.
We observe that the involvement of distance yields an average gain of around 0.2, while imposing hard length constraints degrades the performance.

\begin{figure}
    \centering
    \includegraphics[width=0.32\textwidth]{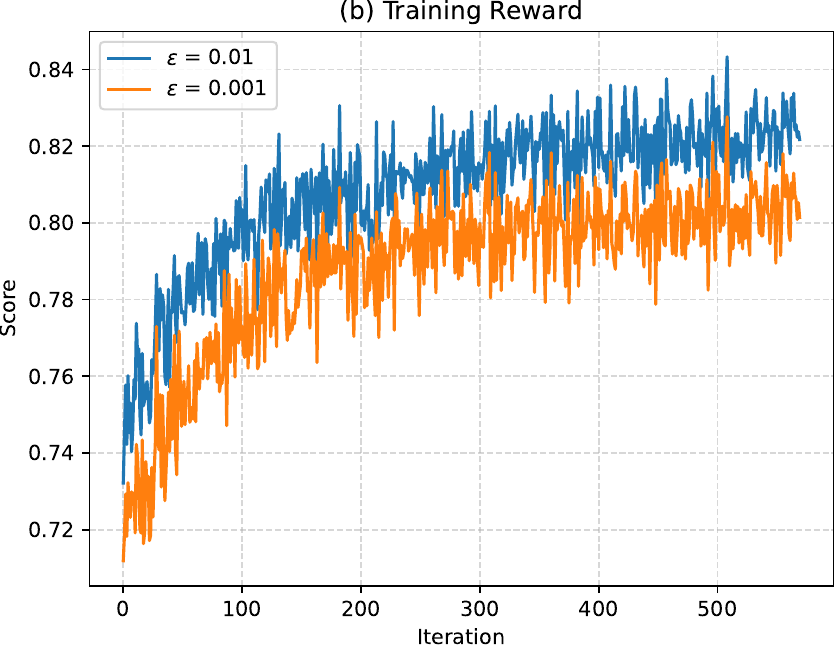}
    \includegraphics[width=0.32\textwidth]{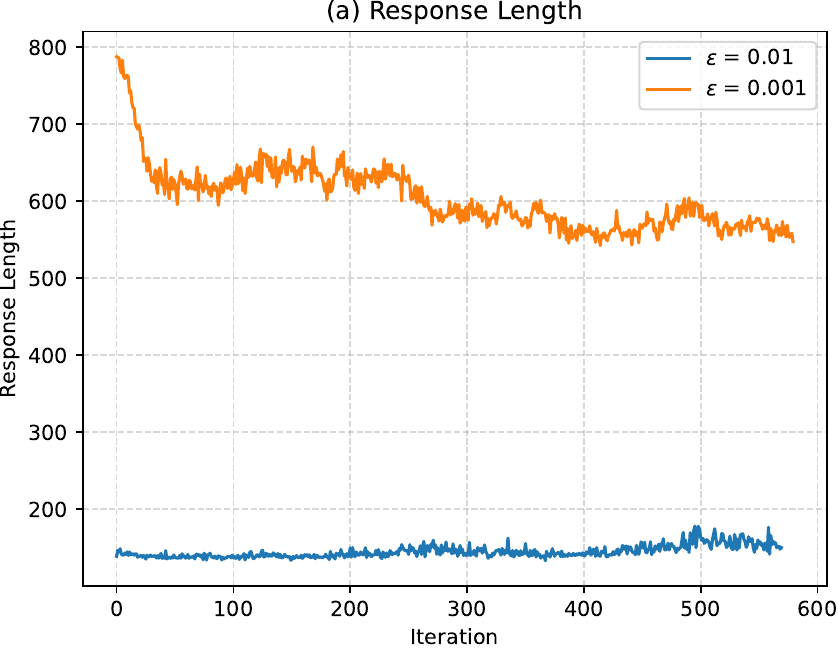}
    \includegraphics[width=0.32\textwidth]{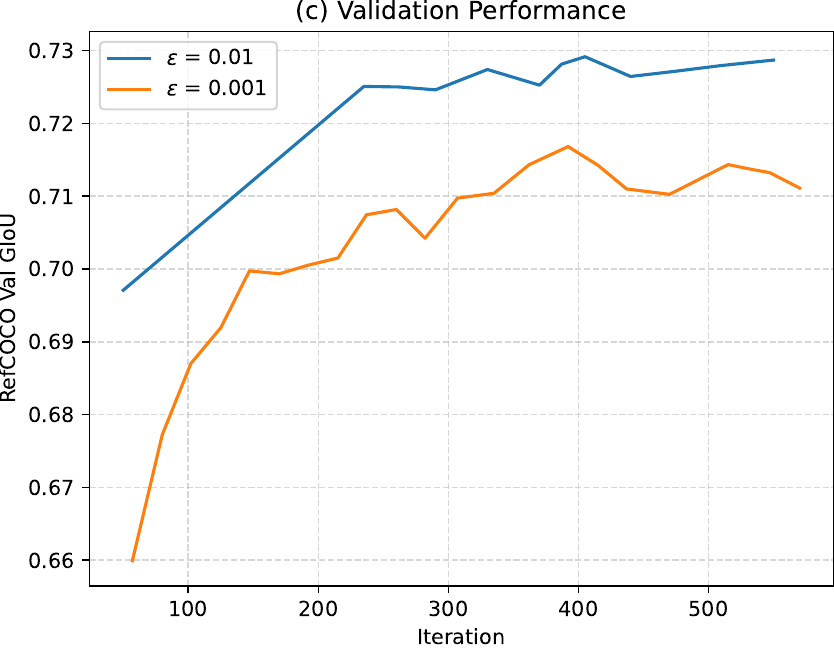}
    \vspace{-2mm}
    \caption{The curve of metrics during the RL stage.}
    \label{fig:rl-curve}
\end{figure}

\paragraph{Metrics Trends During RL}
In \cref{fig:rl-curve}, we illustrate the training states during the reinforcement learning process, including the reward, response length, and validation performance, on different settings of $\epsilon$.
One observation is that, even without length-related rewards, the model can adaptively adjust its output length to keep a reasonable accuracy-efficiency balance during RL.
With a large density of $\epsilon = 0.001$, the number of tokens decreases moderately, trading redundant vertices for token efficiency while preserving mask fidelity.
When the token budget is low with $\epsilon = 0.01$, the response length is increased slightly to better refine the segmentation results.

\begin{figure}
    \centering
    \includegraphics[width=0.9\linewidth]{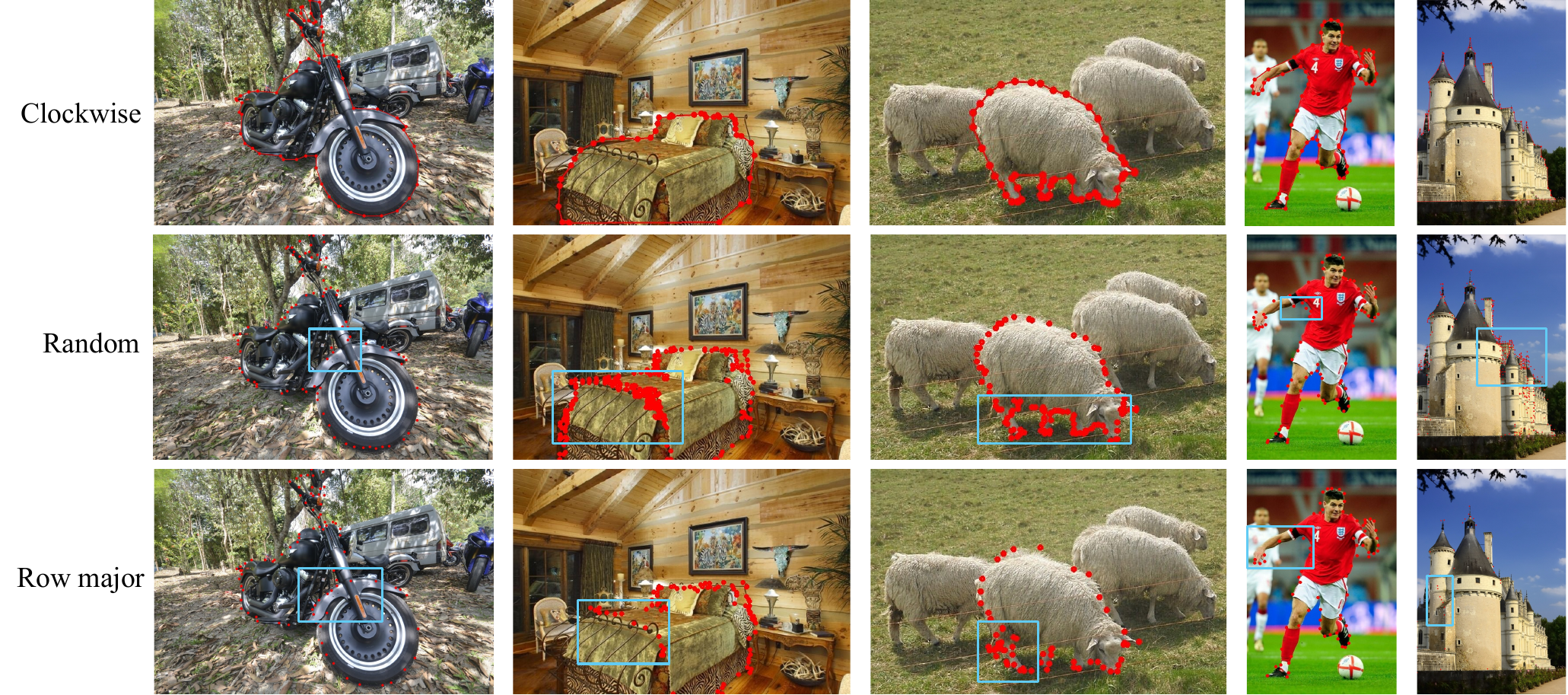}
    \caption{Visual results with different training orders of sampling points. Unsatisfactory prediction points are marked with blue boxes in the image.}
    \vspace{-2mm}
    
    \label{fig:abl-order}
\end{figure}

\paragraph{Order of Sampling Points}
To convert a binary mask to contour coordinates, we apply the Suzuki-Abe algorithm for boundary tracing, which can keep the sampling points in clockwise order.
We conduct experiments with different organizations of sampling points, and display the results in \cref{fig:abl-order}.
First, without enforcing clockwise ordering, the coordinate sequence fails to form a valid polygon, and no segmentation mask can be derived.
Second, clockwise ordering provides a more principled learning target, reducing model entropy.
As shown in the figure, alternative orders confuse the model and yield chaotic or repeated points, decreasing the token efficiency.

\paragraph{Results on extended tasks}
As mentioned in \cref{subsec:data}, we extend the perception task via a unified query interface beyond just the query of the reference phrase.
Namely, our \name~can also achieve the SAM-like functionality, such as $\texttt{point} \rightarrow \texttt{mask}$ and $\texttt{bbox} \rightarrow \texttt{mask}$.
We visualize the results in the Appendix due to limited space.
This significantly demonstrates the generality of our framework, further pushing the upper limit of MLLMs' versatile perception capacity.

\section{Conclusion}
In this work, we demonstrated that a strikingly simple approach—reframing segmentation as the prediction of a sequence of points—is sufficient to unlock a powerful, native capability for pixel-level perception latent within standard MLLM architectures. Our model, \name, cultivated through a novel SFT→RL pipeline, achieves performance that is comparable to, and often surpasses, complex decoder-based systems. It is a finding that high-fidelity perception can be an emergent property of MLLMs. Our work paves the way for a new generation of truly generalist multimodal systems that seamlessly unify perception and reasoning within a single, elegant framework.

\printbibliography[title={References}]

@article{comanici2025gemini,
  title={Gemini 2.5: Pushing the frontier with advanced reasoning, multimodality, long context, and next generation agentic capabilities},
  author={Comanici, Gheorghe and Bieber, Eric and Schaekermann, Mike and Pasupat, Ice and Sachdeva, Noveen and Dhillon, Inderjit and Blistein, Marcel and Ram, Ori and Zhang, Dan and Rosen, Evan and others},
  journal={arXiv preprint arXiv:2507.06261},
  year={2025}
}

@misc{openai2024gpt4ocard,
      title={GPT-4o System Card}, 
      author={OpenAI},
      year={2024},
      eprint={2410.21276},
      archivePrefix={arXiv},
      primaryClass={cs.CL},
      url={https://arxiv.org/abs/2410.21276}, 
}

@article{yin2024survey,
  title={A survey on multimodal large language models},
  author={Yin, Shukang and Fu, Chaoyou and Zhao, Sirui and Li, Ke and Sun, Xing and Xu, Tong and Chen, Enhong},
  journal={National Science Review},
  volume={11},
  number={12},
  pages={nwae403},
  year={2024},
  publisher={Oxford University Press}
}

@article{lu2024deepseek,
  title={Deepseek-vl: towards real-world vision-language understanding},
  author={Lu, Haoyu and Liu, Wen and Zhang, Bo and Wang, Bingxuan and Dong, Kai and Liu, Bo and Sun, Jingxiang and Ren, Tongzheng and Li, Zhuoshu and Sun, Yaofeng and others},
  journal={arXiv preprint arXiv:2403.05525},
  year={2024}
}

@article{Qwen-VL,
  title={Qwen-VL: A Versatile Vision-Language Model for Understanding, Localization, Text Reading, and Beyond},
  author={Bai, Jinze and Bai, Shuai and Yang, Shusheng and Wang, Shijie and Tan, Sinan and Wang, Peng and Lin, Junyang and Zhou, Chang and Zhou, Jingren},
  journal={arXiv preprint arXiv:2308.12966},
  year={2023}
}

@article{liu2023grounding,
  title={Grounding dino: Marrying dino with grounded pre-training for open-set object detection},
  author={Liu, Shilong and Zeng, Zhaoyang and Ren, Tianhe and Li, Feng and Zhang, Hao and Yang, Jie and Li, Chunyuan and Yang, Jianwei and Su, Hang and Zhu, Jun and others},
  journal={arXiv preprint arXiv:2303.05499},
  year={2023}
}

@inproceedings{lai2024lisa,
  title={Lisa: Reasoning segmentation via large language model},
  author={Lai, Xin and Tian, Zhuotao and Chen, Yukang and Li, Yanwei and Yuan, Yuhui and Liu, Shu and Jia, Jiaya},
  booktitle={Proceedings of the IEEE/CVF Conference on Computer Vision and Pattern Recognition},
  pages={9579--9589},
  year={2024}
}

@inproceedings{xia2024gsva,
  title={Gsva: Generalized segmentation via multimodal large language models},
  author={Xia, Zhuofan and Han, Dongchen and Han, Yizeng and Pan, Xuran and Song, Shiji and Huang, Gao},
  booktitle={Proceedings of the IEEE/CVF Conference on Computer Vision and Pattern Recognition},
  pages={3858--3869},
  year={2024}
}

@inproceedings{ren2024pixellm,
  title={Pixellm: Pixel reasoning with large multimodal model},
  author={Ren, Zhongwei and Huang, Zhicheng and Wei, Yunchao and Zhao, Yao and Fu, Dongmei and Feng, Jiashi and Jin, Xiaojie},
  booktitle={Proceedings of the IEEE/CVF Conference on Computer Vision and Pattern Recognition},
  pages={26374--26383},
  year={2024}
}

@inproceedings{he2024multi,
  title={Multi-modal Instruction Tuned LLMs with Fine-grained Visual Perception},
  author={He, Junwen and Wang, Yifan and Wang, Lijun and Lu, Huchuan and He, Jun-Yan and Lan, Jin-Peng and Luo, Bin and Xie, Xuansong},
  booktitle={Proceedings of the IEEE/CVF Conference on Computer Vision and Pattern Recognition},
  pages={13980--13990},
  year={2024}
}

@inproceedings{zhang2024groundhog,
  title={Groundhog: Grounding large language models to holistic segmentation},
  author={Zhang, Yichi and Ma, Ziqiao and Gao, Xiaofeng and Shakiah, Suhaila and Gao, Qiaozi and Chai, Joyce},
  booktitle={Proceedings of the IEEE/CVF conference on computer vision and pattern recognition},
  pages={14227--14238},
  year={2024}
}

@inproceedings{rasheed2024glamm,
  title={Glamm: Pixel grounding large multimodal model},
  author={Rasheed, Hanoona and Maaz, Muhammad and Shaji, Sahal and Shaker, Abdelrahman and Khan, Salman and Cholakkal, Hisham and Anwer, Rao M and Xing, Eric and Yang, Ming-Hsuan and Khan, Fahad S},
  booktitle={Proceedings of the IEEE/CVF Conference on Computer Vision and Pattern Recognition},
  pages={13009--13018},
  year={2024}
}

@article{zhang2023next,
  title={Next-chat: An lmm for chat, detection and segmentation},
  author={Zhang, Ao and Zhao, Liming and Xie, Chen-Wei and Zheng, Yun and Ji, Wei and Chua, Tat-Seng},
  journal={arXiv preprint arXiv:2311.04498},
  year={2023}
}

@inproceedings{mao2016generation,
  title={Generation and comprehension of unambiguous object descriptions},
  author={Mao, Junhua and Huang, Jonathan and Toshev, Alexander and Camburu, Oana and Yuille, Alan L and Murphy, Kevin},
  booktitle={Proceedings of the IEEE conference on computer vision and pattern recognition},
  pages={11--20},
  year={2016}
}

@inproceedings{kazemzadeh2014referitgame,
  title={Referitgame: Referring to objects in photographs of natural scenes},
  author={Kazemzadeh, Sahar and Ordonez, Vicente and Matten, Mark and Berg, Tamara},
  booktitle={Proceedings of the 2014 conference on empirical methods in natural language processing (EMNLP)},
  pages={787--798},
  year={2014}
}

@article{wang2024visionllm,
  title={Visionllm: Large language model is also an open-ended decoder for vision-centric tasks},
  author={Wang, Wenhai and Chen, Zhe and Chen, Xiaokang and Wu, Jiannan and Zhu, Xizhou and Zeng, Gang and Luo, Ping and Lu, Tong and Zhou, Jie and Qiao, Yu and others},
  journal={Advances in Neural Information Processing Systems},
  volume={36},
  year={2024}
}

@article{wu2024visionllmv2,
  title={VisionLLM v2: An End-to-End Generalist Multimodal Large Language Model for Hundreds of Vision-Language Tasks},
  author={Jiannan, Wu and Muyan, Zhong and Sen, Xing and Zeqiang, Lai and Zhaoyang, Liu and Zhe, Chen and Wenhai, Wang and Xizhou, Zhu and Lewei, Lu and Tong, Lu and Ping, Luo and Yu, Qiao and Jifeng, Dai},
  journal={arXiv preprint arXiv:2406.08394},
  year={2024}
}

@article{wu2024towards,
  title={Towards Semantic Equivalence of Tokenization in Multimodal LLM},
  author={Wu, Shengqiong and Fei, Hao and Li, Xiangtai and Ji, Jiayi and Zhang, Hanwang and Chua, Tat-Seng and Yan, Shuicheng},
  journal={arXiv preprint arXiv:2406.05127},
  year={2024}
}

@misc{liu2024llava,
  title={Llava-next: Improved reasoning, ocr, and world knowledge},
  author={Liu, Haotian and Li, Chunyuan and Li, Yuheng and Li, Bo and Zhang, Yuanhan and Shen, Sheng and Lee, Yong Jae},
  year={2024}
}

@article{zhang2024omg,
  title={Omg-llava: Bridging image-level, object-level, pixel-level reasoning and understanding},
  author={Zhang, Tao and Li, Xiangtai and Fei, Hao and Yuan, Haobo and Wu, Shengqiong and Ji, Shunping and Loy, Chen Change and Yan, Shuicheng},
  journal={arXiv preprint arXiv:2406.19389},
  year={2024}
}

@article{chen2021pix2seq,
  title={Pix2seq: A language modeling framework for object detection},
  author={Chen, Ting and Saxena, Saurabh and Li, Lala and Fleet, David J and Hinton, Geoffrey},
  journal={arXiv preprint arXiv:2109.10852},
  year={2021}
}

@article{wei2024lasagna,
  title={LaSagnA: Language-based Segmentation Assistant for Complex Queries},
  author={Wei, Cong and Tan, Haoxian and Zhong, Yujie and Yang, Yujiu and Ma, Lin},
  journal={arXiv preprint arXiv:2404.08506},
  year={2024}
}

@article{lan2024text4seg,
  title={Text4seg: Reimagining image segmentation as text generation},
  author={Lan, Mengcheng and Chen, Chaofeng and Zhou, Yue and Xu, Jiaxing and Ke, Yiping and Wang, Xinjiang and Feng, Litong and Zhang, Wayne},
  journal={arXiv preprint arXiv:2410.09855},
  year={2024}
}

@article{tang2025ufo,
  title={Ufo: A unified approach to fine-grained visual perception via open-ended language interface},
  author={Tang, Hao and Xie, Chenwei and Wang, Haiyang and Bao, Xiaoyi and Weng, Tingyu and Li, Pandeng and Zheng, Yun and Wang, Liwei},
  journal={arXiv preprint arXiv:2503.01342},
  year={2025}
}

@article{shi2024seededit,
  title={Seededit: Align image re-generation to image editing},
  author={Shi, Yichun and Wang, Peng and Huang, Weilin},
  journal={arXiv preprint arXiv:2411.06686},
  year={2024}
}

@inproceedings{wang2025mllm,
  title={Mllm-tool: A multimodal large language model for tool agent learning},
  author={Wang, Chenyu and Luo, Weixin and Dong, Sixun and Xuan, Xiaohua and Li, Zhengxin and Ma, Lin and Gao, Shenghua},
  booktitle={2025 IEEE/CVF Winter Conference on Applications of Computer Vision (WACV)},
  pages={6678--6687},
  year={2025},
  organization={IEEE}
}

@article{liu2025infiguiagent,
  title={Infiguiagent: A multimodal generalist gui agent with native reasoning and reflection},
  author={Liu, Yuhang and Li, Pengxiang and Wei, Zishu and Xie, Congkai and Hu, Xueyu and Xu, Xinchen and Zhang, Shengyu and Han, Xiaotian and Yang, Hongxia and Wu, Fei},
  journal={arXiv preprint arXiv:2501.04575},
  year={2025}
}

@article{qin2025ui,
  title={Ui-tars: Pioneering automated gui interaction with native agents},
  author={Qin, Yujia and Ye, Yining and Fang, Junjie and Wang, Haoming and Liang, Shihao and Tian, Shizuo and Zhang, Junda and Li, Jiahao and Li, Yunxin and Huang, Shijue and others},
  journal={arXiv preprint arXiv:2501.12326},
  year={2025}
}

@article{team2025kimivl,
  title={Kimi-vl technical report},
  author={Team, Kimi and Du, Angang and Yin, Bohong and Xing, Bowei and Qu, Bowen and Wang, Bowen and Chen, Cheng and Zhang, Chenlin and Du, Chenzhuang and Wei, Chu and others},
  journal={arXiv preprint arXiv:2504.07491},
  year={2025}
}

@article{liu2025muon,
  title={Muon is scalable for LLM training},
  author={Liu, Jingyuan and Su, Jianlin and Yao, Xingcheng and Jiang, Zhejun and Lai, Guokun and Du, Yulun and Qin, Yidao and Xu, Weixin and Lu, Enzhe and Yan, Junjie and others},
  journal={arXiv preprint arXiv:2502.16982},
  year={2025}
}

@article{zheng2025group,
  title={Group sequence policy optimization},
  author={Zheng, Chujie and Liu, Shixuan and Li, Mingze and Chen, Xiong-Hui and Yu, Bowen and Gao, Chang and Dang, Kai and Liu, Yuqiong and Men, Rui and Yang, An and others},
  journal={arXiv preprint arXiv:2507.18071},
  year={2025}
}

@article{shao2024deepseekmath,
  title={Deepseekmath: Pushing the limits of mathematical reasoning in open language models},
  author={Shao, Zhihong and Wang, Peiyi and Zhu, Qihao and Xu, Runxin and Song, Junxiao and Bi, Xiao and Zhang, Haowei and Zhang, Mingchuan and Li, YK and Wu, Yang and others},
  journal={arXiv preprint arXiv:2402.03300},
  year={2024}
}

@article{guo2025deepseek,
  title={Deepseek-r1: Incentivizing reasoning capability in llms via reinforcement learning},
  author={Guo, Daya and Yang, Dejian and Zhang, Haowei and Song, Junxiao and Zhang, Ruoyu and Xu, Runxin and Zhu, Qihao and Ma, Shirong and Wang, Peiyi and Bi, Xiao and others},
  journal={arXiv preprint arXiv:2501.12948},
  year={2025}
}

@article{team2025kimik15,
  title={Kimi k1. 5: Scaling reinforcement learning with llms},
  author={Team, Kimi and Du, Angang and Gao, Bofei and Xing, Bowei and Jiang, Changjiu and Chen, Cheng and Li, Cheng and Xiao, Chenjun and Du, Chenzhuang and Liao, Chonghua and others},
  journal={arXiv preprint arXiv:2501.12599},
  year={2025}
}

@article{suzuki1985topological,
  title={Topological structural analysis of digitized binary images by border following},
  author={Suzuki, Satoshi and others},
  journal={Computer vision, graphics, and image processing},
  volume={30},
  number={1},
  pages={32--46},
  year={1985},
  publisher={Elsevier}
}

@misc{itseez2015opencv,
  title={Open Source Computer Vision Library},
  author={Itseez},
  year={2015},
  howpublished = {\url{https://github.com/itseez/opencv}}
}

@inproceedings{wang2024git,
  title={Git: Towards generalist vision transformer through universal language interface},
  author={Wang, Haiyang and Tang, Hao and Jiang, Li and Shi, Shaoshuai and Naeem, Muhammad Ferjad and Li, Hongsheng and Schiele, Bernt and Wang, Liwei},
  booktitle={European Conference on Computer Vision},
  pages={55--73},
  year={2024},
  organization={Springer}
}

@article{chen2022unified,
  title={A unified sequence interface for vision tasks},
  author={Chen, Ting and Saxena, Saurabh and Li, Lala and Lin, Tsung-Yi and Fleet, David J and Hinton, Geoffrey E},
  journal={Advances in Neural Information Processing Systems},
  volume={35},
  pages={31333--31346},
  year={2022}
}

\newpage
\appendix

\section{Limitations and Diagnostics}
While \name~eliminates task decoders, long sequences remain a bottleneck for high-resolution, highly-curved objects.
Errors tend to cluster at sharp corners and thin structures under aggressive sparsification.
Future diagnostics should consider boundary F-score, vertex-wise Chamfer distance, and token-per-mask analyses across object scales to complement cIoU/Acc@0.5.

\section{Additional Implementation Details}

\subsection{Data Details}

\textbf{Pre-training Data}: We leveraged large-scale open-source and web data, including LAION and Coyo. All pre-training samples were labeled using the annotation pipeline detailed in \cref{subsec:data}.

\textbf{SFT and RL Data}: For Supervised Fine-Tuning (SFT), we utilized the RefCOCO series, strictly adhering to the data processing protocol established in Text4Seg~\cite{lan2024text4seg}.
Specifically, we construct the training dataset with the \texttt{train} split of refCOCO, refCOCO+~\cite{kazemzadeh2014referitgame}, refCOCOg~\cite{mao2016generation}, and refCLEF, resulting in a dataset of 800k samples.
Similarly, the prompt set with 400k samples for Reinforcement Learning (RL) was also derived from the RefCOCO series.

\textbf{Note on Benchmarking}: It is important to note that the results presented in \cref{tab:results_res} and \cref{tab:results_rec} are intended to ensure a fair comparison with state-of-the-art methods. Therefore, these reported metrics are derived from models trained exclusively with the SFT and RL stages (i.e., with only RefCOCO datasets). The discussion regarding pre-training with large-scale web data was included primarily as a scaling analysis and ablative study, as demonstrated in \cref{tab:abl-stage}.

\subsection{Training Details}

\cref{tab:params-sft} and \cref{tab:param-rl} present the training hyper-parameters used in SFT and RL stages.

\begin{table}[!h]
    \centering
    \begin{minipage}[t]{0.49\textwidth}
        \centering
        \caption{Training settings for SFT stage.}
        \label{tab:params-sft}
        \begin{tabular}{c|l|c}
            \toprule
             & Param Name & Value \\
            \midrule
            \multirow{8}{*}{Optimizer} & Type  & Enhanced Muon  \\
             & Max Learning rate   & 5e-5 \\
             & Min Learning rate   & 2e-6 \\ 
             & Weight decay  & 0.1 \\
             & $(\beta_1, \beta_2)$  & (0.9, 0.95) \\
             & Gradient norm clip    & 1.0 \\ 
             & Scheduler & Cosine decay \\
             & Warmup ratio & 0.03 \\
            \midrule
            \multirow{4}{*}{Training}  & Numerical precision   &  FP16 \\
             & Global batch size     &  256 \\
             & Samples per epoch & 800k \\ 
             & Total epochs          &  1 \\
            \bottomrule
        \end{tabular}
    \end{minipage}
    \hfill 
    \begin{minipage}[t]{0.49\textwidth}
        \centering
        \caption{Training settings for RL stage.}
        \label{tab:param-rl}
        \begin{tabular}{c|l|c}
            \toprule
             & Param Name & Value \\
            \midrule
            \multirow{4}{*}{Algorithm} & Type & GSPO \\
             & Clip Ratio & [3e-4, 4e-4] \\
             & KL Alpha & 0.1 \\
             & Responses / Group & 8 \\ 
            \midrule
            \multirow{4}{*}{Optimizer} & Type  & Enhanced Muon  \\
             & Constant LR   & 2e-6 \\
             & $(\beta_1, \beta_2)$  & (0.9, 0.95) \\
             & Gradient norm clip    & 1.0 \\ 
            \midrule
            \multirow{5}{*}{Training}  & Numerical precision   &  FP16 \\
             & Rollout Temp    & 0.8 \\
             & Global batch size     &  256 \\
             & Samples per epoch & 400k \\
             & Total epochs          &  2 \\
            \bottomrule
        \end{tabular}
    \end{minipage}
\end{table}

\section{Additional qualitative results}
\label{sec:more-results}
\begin{figure}[t]
    \centering
    \includegraphics[width=0.9\linewidth]{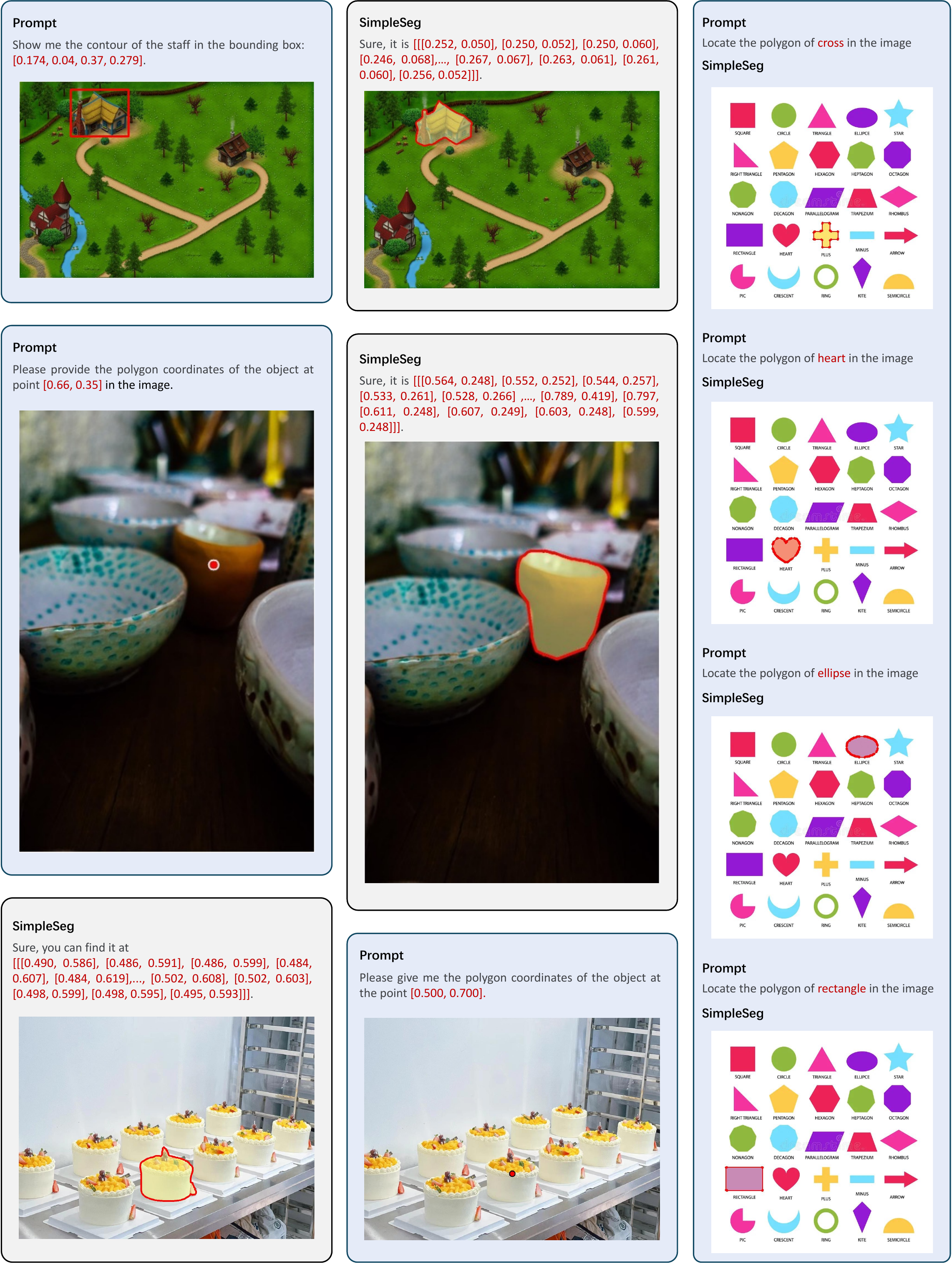}
    \caption{More results on more diverse tasks, including \texttt{(point$\to$mask)} and \texttt{(bbox$\to$mask)}. The position information is visualized in the image.}
    \label{fig:more1}
\end{figure}

\begin{figure}[t]
    \centering
    \includegraphics[width=0.85\linewidth]{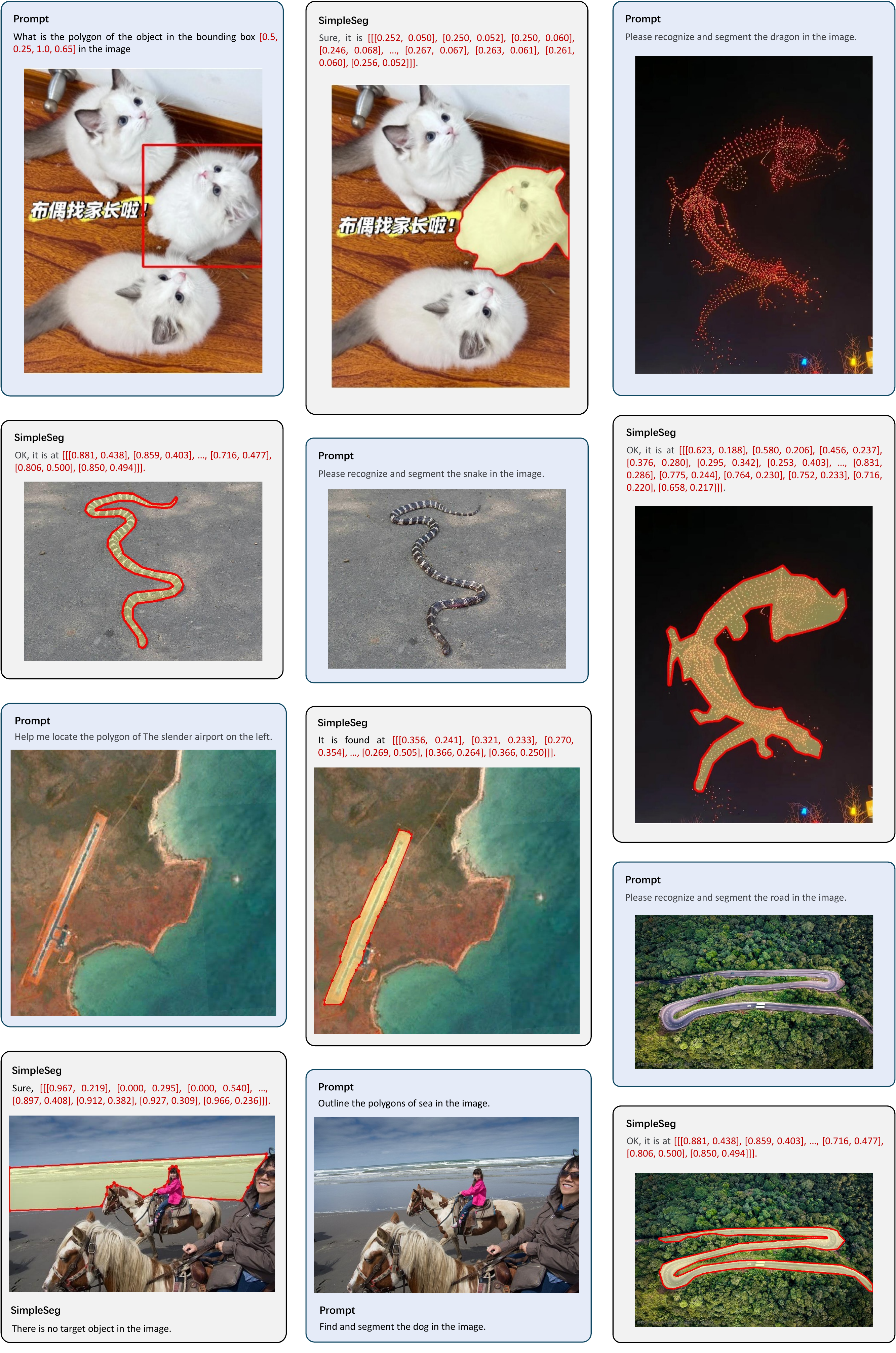}
    \caption{More results on more diverse tasks including \texttt{(bbox$\to$mask)} and \texttt{(text$\to$mask)}. The position information is visualized in the image.}
    \label{fig:more2}
\end{figure}

\begin{figure}[t]
    \centering
    \includegraphics[width=0.82\linewidth]{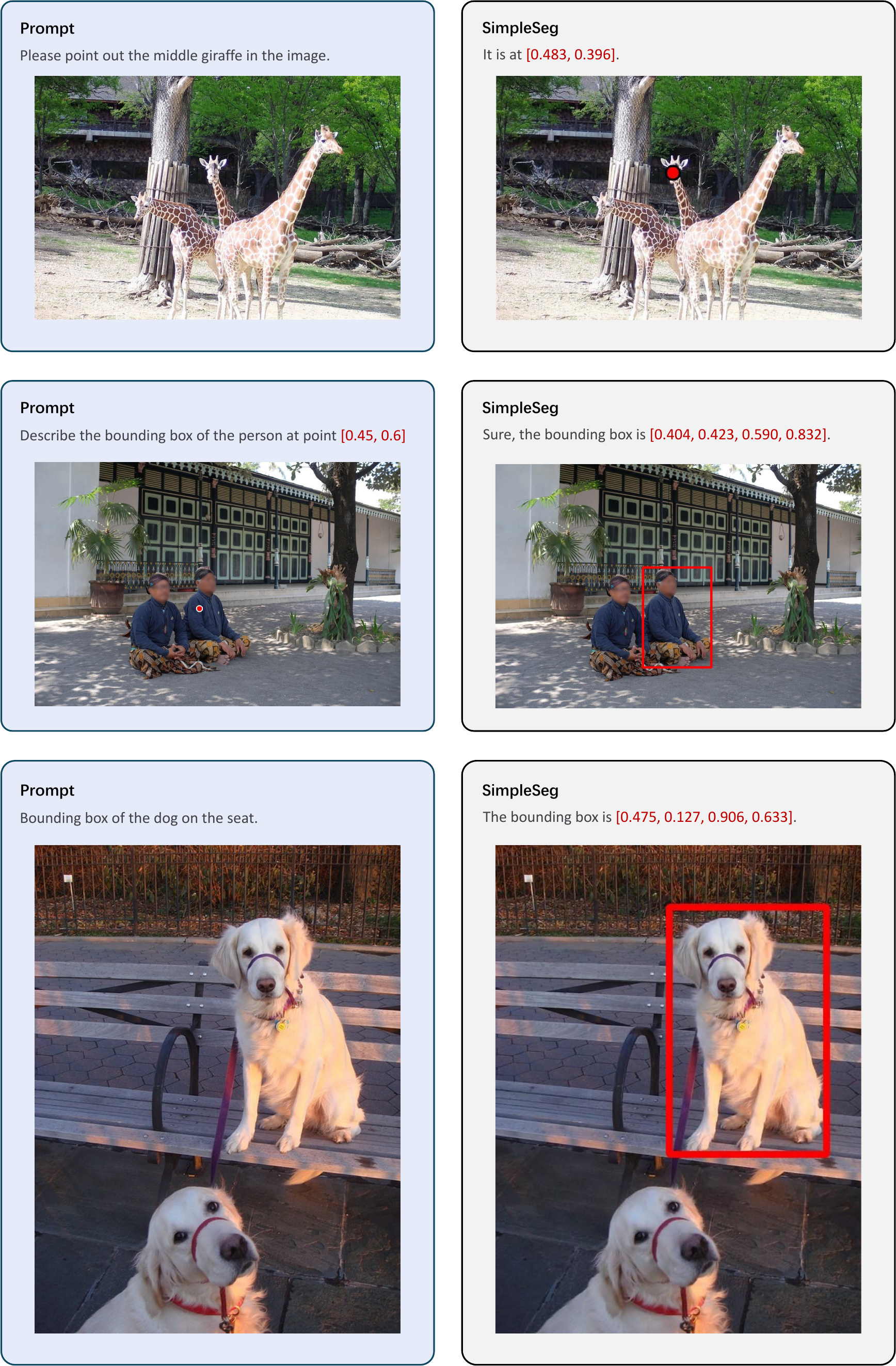}
    \caption{More results on more diverse tasks, such as \texttt{(text$\to$point)} and \texttt{(text$\to$bbox)}. The position information is visualized in the image.}
    \label{fig:more3}
\end{figure}

As shown in \cref{fig:more1}, \cref{fig:more2}, and \cref{fig:more3}, we provide more example results of \name~on our extended tasks, which significantly demonstrate our \name's accuracy, robustness, and generalization.

\begin{figure}[t]
    \centering
    \includegraphics[width=0.8\linewidth]{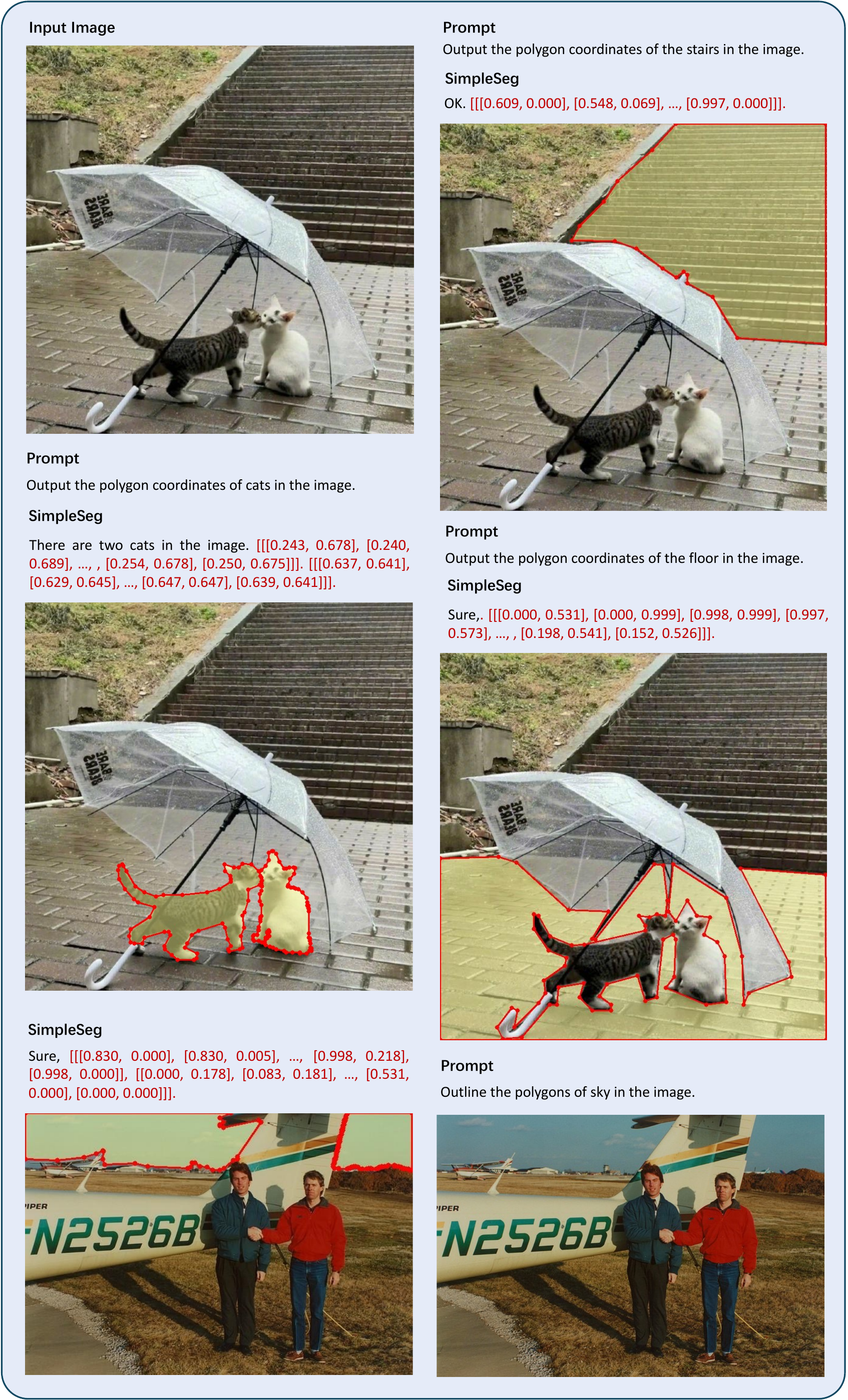}
    \caption{More results of panoptic segmentation, multiple objects segmentation, and multi-parts object segmentation.}
    \label{fig:more4}
\end{figure}

\begin{figure}[t]
    \centering
    \includegraphics[width=0.85\linewidth]{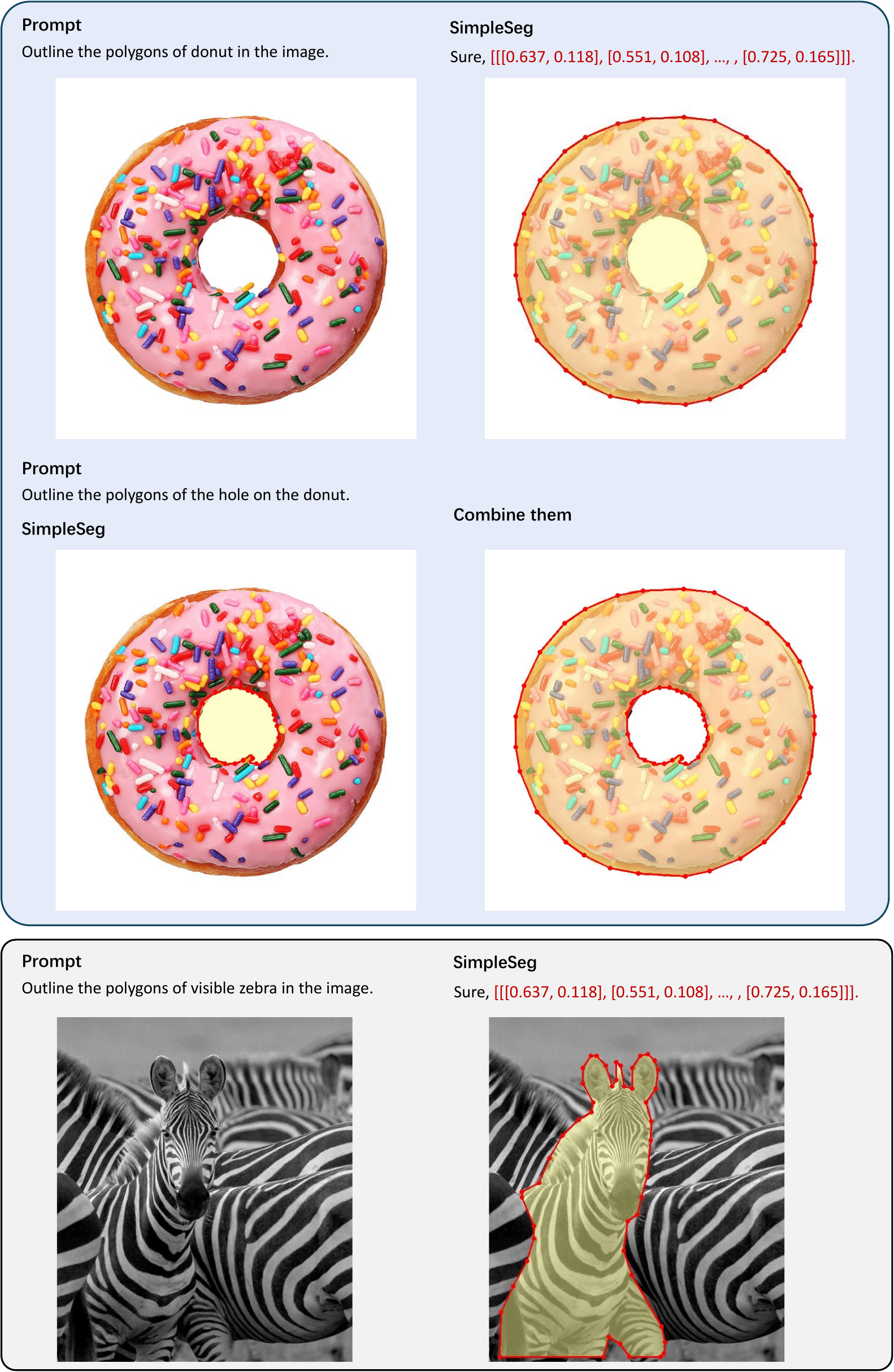}
    \caption{Visual results of failure cases, e.g., object with holes and texture confusion.}
    \label{fig:more5}
\end{figure}

\end{document}